# COVID-19 Spreading Prediction and Impact Analysis by using Artificial Intelligence for Sustainable Global Health Assessment

Subhrangshu Adhikary[1], Sonam Chaturvedi[2], Sudhir Kumar Chaturvedi[3,*], and Saikat Banerjee[4]

[1]Department of Computer Science and Engineering, Dr. B.C. Roy Engineering College, Durgapur-713206, West Bengal, India
subhrangshu.adhikary@spiraldevs.com
[2]Department of Health Safety Environment and Civil Engineering, University of Petroleum and Energy Studies (UPES), Bidholi Energy Acres, Dehradun 248007, India
sonam.dobriyal@gmail.com
[3,*]Department of Aerospace Engineering, UPES, Dehradun-248007, India
sudhir.chaturvedi@ddn.upes.ac.in
[4]Department of Mechanical engineering, Cubicx, Kolkata-700070, West Bengal, India
*saikatbanerjee@cubicxindia.com*

**Abstract.** The COVID-19 pandemic is considered as the most alarming global health calamity of this century. COVID-19 has been confirmed to be mutated from coronavirus family. As stated by the records of The World Health Organization (WHO at April 18 2020), the present epidemic of COVID-19, has influenced more than 2,164,111 persons and killed more than 146,198 folks in over 200 countries across the globe and billions had confronted impacts in lifestyle because of this virus outbreak. The ongoing overall outbreak of the COVID-19 opened up new difficulties to the research sectors. Artificial intelligence (AI) driven strategies can be valuable to predict the parameters, hazards, and impacts of such an epidemic in a cost-efficient manner. The fundamental difficulties of AI in this situation is the limited availability of information and the uncertain nature of the disease. Here in this article, we have tried to integrate AI to predict the infection outbreak and along with this, we have also tried to test whether AI with help deep learning can recognize COVID-19 infected chest X-Rays or not. The global outbreak of the virus posed enormous economic, ecological and societal challenges into the human population and with help of this paper, we have tried to give a message that AI can help us to identify certain features of the disease outbreak that could prove to be essential to protect the humanity from this deadly disease.

## 1 Introduction

The widespread outbreak of COVID-19 from the city of Wuhan, China has spread all over the world has affected humanity drastically. Millions of people have been infected by the virus and over five hundred thousands of people have already died because of the virus. The remaining people all over the world is at risk of contamination. COVID-19 is a single-stranded RNA virus from the coronavirus family. It has spiked glycoproteins around its body surface which attach to receptor cells of the host and infects primarily lungs causing auto-immune response around alveoli which in turn swells the surrounding area and lungs filled up with fluids. This makes the gaseous exchange of the lungs difficult and severely infected patients require a life support system to continue a sufficient oxygen supply. There is no specific cure available to fight against the virus and only generic treatments are being used to reduce the damage caused by the infection. Researchers are working hard to discover a vaccine or antiviral treatment to fight against the disease. Until then, prevention of the disease from spreading is the best strategy to minimize the contamination.

Artificial Intelligence (AI) can be a very useful tool to control the pandemic situation. It could be used in many situations like aiding researchers to simulate the vaccination process, aid doctors to monitor patients and alert doctors if a patient requires special care so doctors can attend as many patients as possible, aid shopping malls and markets to monitor the crowd and enforce social distancing measures, many more. Usage of AI in all these segments could help to optimize the pandemic control capacity in an inexpensive and fast deployable approach. As AI is a software program which can be transmitted quickly and can work with a wide variety of infrastructure setup available worldwide, it can be deployed rapidly. However, the uncertain nature of the pandemic and limited availability of quality data to train on is the biggest hurdle for AI. However, implementations of advanced techniques like usage of generative algorithm and transferring of pre-trained weights for deep learning can be very useful to train and predict outcomes with high accuracy even with a very limited dataset. In this paper, we have demonstrated the use of AI in two cases.

First, we have used generative machine learning algorithm with Bayesian Ridge regressor technique which works on a probabilistic approach to reconstruct a wide range of data with given inputs and Gaussian Ridge regressor technique which is an extension of the earlier. By these, we have built a machine learning model which can predict the upcoming future trends of the virus based on currently available data so that early measures could be taken to prevent as much contamination as possible. The model trains on two sets, to estimate the amount of confirmed COVID-19 cases and to estimate the number of death cases going to take place in the near future. The model is described in section 3.2 and its performances are discussed in section 4.1.

Second, we have used deep learning on chest X-Ray images to classify COVID-19 infected patients from normal healthy lungs as well as lungs infected by other lung diseases like Streptococcus pneumonia infection, Acute Respiratory Distress Syndrome (ARDS), etc. COVID-19 causes swelling in the lungs which are spotted in X-Rays. Normal lungs would not have such swelling and should be easily classified by the model. However, other lung infections as stated above also form swelling and fluid build-ups in their unique pattern. The model that we have built, has analysed all these features and successfully classified the two classes. The model is described in section 3.3 and its performances are discussed in section 4.2.

The paper is arranged in the following manner. The recent advancements in studies related to COVID-19 and AI are listed in section 2, section 3 contains the availability of the dataset and its preprocessing, procedure of making our models, section 4 contains the results observed from our experiment and discussed their performances and section 5 consist of the conclusions we have drawn from the experiment.

**2 Background Works**
Here in this section, we have discussed the origin, evolution and biological background of COVID-19 along with usages of AI for controlling the pandemic, specifically the usage of machine learning and deep learning for the same purpose.

**2.1 Background of COVID-19 pandemic**
The epidemiology and pathogenesis of COVID-19 have been discussed in [1]. In this, the phylogenetic analysis of the virus revealed that it is potentially a zoonotic virus which means it spreads between animal and people and patients develop intestinal symptoms like diarrhoea which was rarely evident in the case of MERS-CoV and SARS-CoV. A summary report based on 72314 cases from the Chinese center for disease control and prevention has been made which shows the epidemiological characteristics of the disease [2]. A summarized report on the origin, transmission medium and clinical therapies on COVID-19 have been specified in [3]. The utilization of critical care for the COVID-19 outbreak in Lombardy, Italy for an emergency response to the situation is described in [4]. Migrations have been one of the major reason for the virus outbreak in different places and the effect of travel restriction on the spread of the virus is discussed in [5]. The social distancing method to avoid human to human contact is an effective strategy to control the virus outbreak and the feasibility of this has been discussed in [6]. The demographic science was used to understand the spread and fatality rate caused by the virus [7].

COVID-19 is a single-stranded RNA virus and the structure of RNA-dependent polymerase of the virus is discussed in [8]. The virus has a round body with a spiked glycoprotein structure which attaches itself to the receptors of host cells. The glycan shield and the spikes' interaction with human have been predicted in [9,10]. Genotype is the genetic characteristic of a cell and phenotype is the corresponding physical characteristic of the cell. Genotypic and phenotypic characterization of the COVID-19 cell and their roles in pathogenesis was discussed in [11]. The virology epidemiology, pathogenesis and the control measures for the pandemic have been discussed in details in [12].

The Food and Drug Administration (FDA) assures the safety and effectiveness of drugs for public use, the virtual screening of FDA approved drugs to give relief to COVID-19 patients were conducted by authors and discussed in [13]. The use of chloroquine phosphate for the treatment of COVID-19 associated pneumonia has shown positive results [14]. The usage of randomized drugs on clinical trials and their outcomes were studied and are recorded in [15]. COVID-19 causes inflammation within the lungs with the build-up of fluids. Therefore the usage of anti-inflammatory drugs for the treatment of COVID-19 patients with the severe condition can give relief for breathing. This clinical immunologists from China have provided their perspective about the matter in [16]. Anti-inflammatory drugs are often steroidal, but to study the effects of non-steroidal anti-inflammatory drugs on COVID-19 patients, experiments were conducted and the outcomes were recorded in [17].

## 2.2 Usage of AI for controlling the COVID-19 pandemic

Since the evolution of better processing capacity of computers, AI has become a very essential tool to assist human in multiple aspects. AI has intervened the processes where human intervention is not practically feasible or is expensive. AI has the decision-making capabilities which can be primarily in the form of classification and regression. AI was earlier used to form a framework for quicker identification of COVID-19 in cities and towns under quarantine with a mobile phone-based survey model [18]. Chest CT scan reports were used to distinguish COVID-19 from community-acquired pneumonia with AI [19]. The authors in [20] have reviewed the use of AI to fight against COVID-19 and given prominent perspective about the matter based on all recent studies. AI was coupled with universal data sharing standards to monitor human health in smart city networks [21].

## 3 Methodology

Here we have discussed the methods we have implemented to test different artificial intelligence models which can be useful for controlling the spread of the COVID-19 outbreak. This section is divided into two primary sub-sections. One discusses the use of machine learning regression model to predict the trend of the virus outbreak, that is the prediction of future deaths and confirmation of COVID-19 positive cases based on current data. This is based on two generative machine learning regressors known as Bayesian Ridge Regressor and Gaussian Process Regressor. The other section discusses the implementation of deep learning technology for the detection of COVID-19 positive patients with the help of chest X-Rays as an alternative to traditional lab test. This is performed by deep convolution neural network model. Details of the procedures for both are given in following sub-section. The performances of all these techniques are recorded in section 4.

### 3.1 Data availability and pre-processing

The regression model for estimating future confirmed cases and death cases have been made with the help of publicly available records provided by Johns Hopkins University [22]. For the X-Ray, classification has used Kaggle dataset [23,24] which were cited by many peer-reviewed articles. These datasets combined contained over four thousand image data for chest X-Ray including normal subjects, COVID-19 patients, pneumonia patients, ARDS patients and more. The datasets, however, contained some noise in the form of CT-Scan images which we have manually removed. Later we have split them into two sets for training and testing purpose.

### 3.2 Virus outbreak trend prediction with machine learning

The trend prediction of the virus is to estimate confirmed cases and deaths in future based on present data we have used two popular generative machine learning regressor known as Bayesian Ridge Regressor and Gaussian Process Regressor.

**Bayesian Ridge Regressor (BRR)**

Bayesian Ridge Regressor is a generative regressor which can reconstruct missing data or poorly distributed data by creating a linear regression model with the help of probability distributors instead of estimating from available points for training. As the number of deaths and confirmed cases will grow with time and will not oscillate within a specific range, a reconstruction of points from available points are necessary and BRR algorithms fit perfectly with this situation. Based on currently available data, it builds a probabilistic model estimating the change in the dependent feature (future trend) with respect to the independent feature (past trend) and this probability measure is used to estimate newer sets of data.

**Gaussian Process Regressor (GPR)**

GPR is a non-parametric regressor which works on the principle of the probability distribution for all functions that fit the data instead of calculating the probability distribution of parameters of a specific function. GPR can train with a very small dataset and provide good outcomes from them. It works like a multivariate Gaussian distribution of infinite-dimension and the space of functions could be incorporated by the selection of the mean and covariance function. A GPR has multiple kernels and our regressor is primarily based on two kernels, dot product kernel and white kernel. A dot product kernel can be obtained from linear regression by tuning the coefficients and biases. It is invariant to coordinate rotation about the origin but not translation. A white kernel, on the other hand, is a part of sum kernel which explains the noise of the signal as an independent and identical normal distribution.

### 3.3 Detection of COVID-19 from chest X-Rays with deep learning

Deep learning has been evolved as a powerful tool for classification and regression problem for both supervised and unsupervised learning. The deep convolution neural network (CNN) is the most promising algorithm to work with image data. A CNN traverses through all images in a dataset and extracts several features from it like the colour cluster concentration, texture, boundary, etc. out of which the most contrasting features are filtered and establish a classifiable relationship between the different classes or categories of the image datasets. We have trained the classifiers to figure out certain contrasting features between two classes of data, that is, one dataset containing chest X-Ray of patients infected with COVID-19 and other dataset contains a combination of normal chest X-Rays and other lung infection apart from COVID-19. The performances are recorded in section 4.2.

## 4 Results and Discussions

The performance of the outbreak trend predictor model is discussed in section 4.1 and performance analysis of COVID-19 detection with chest X-Rays model is discussed in section 4.2.

### 4.1 Performance analysis of outbreak trend predictor model

The daily recorded values of total confirmed cases and death cases of India were recorded from the first detected confirmed case up to the next 100 days and these data were split into two parts. The records of confirmed cases were split and first 67% of the data were used to train the model and the other 33% of the data were used to validate the accuracy of the model. Due to the uncertain nature of the disease, we needed more data to train the model to predict the future trend of death case and because of this, we have used 75% of the death record data for training and 25% of the data for validation purpose. To estimate the performance of the data, we have used Root Mean Squared Error (RMS Error) metric along with the lower and upper bound of the testing dataset and also the training and testing time for each model. The performances are recorded in table 1 and the graph of the prediction is plotted in fig. 1 and fig. 2.

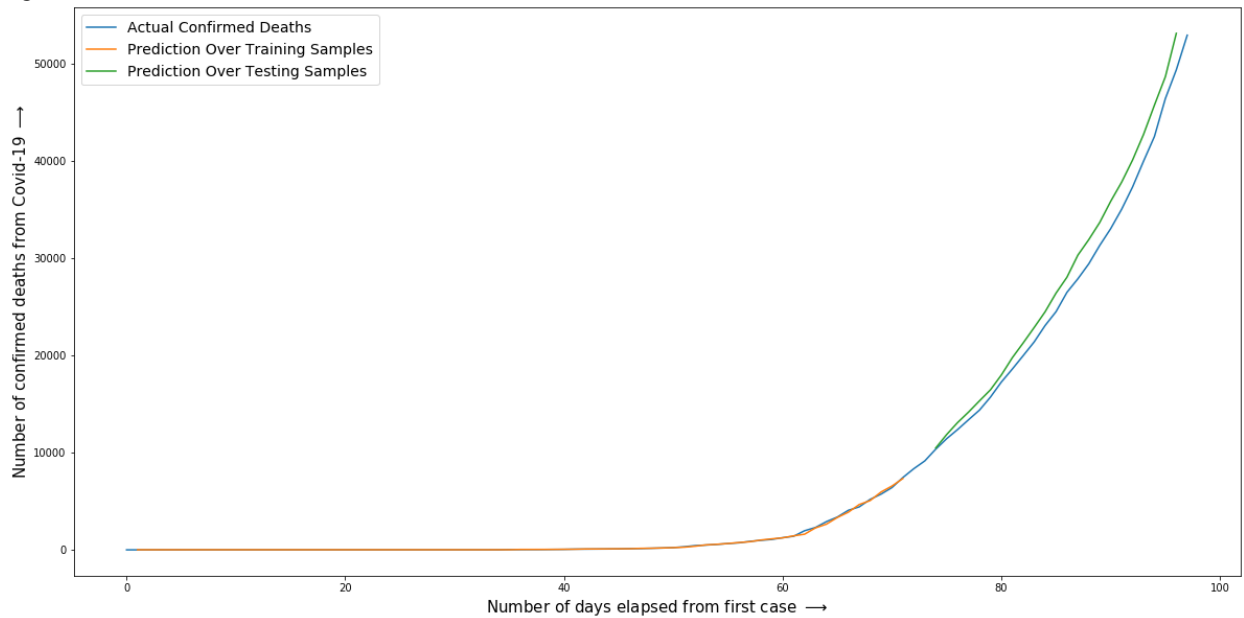

**Fig. 1.** The predicted trend of COVID-19 confirmed cases from 1st confirmed case till 100 days

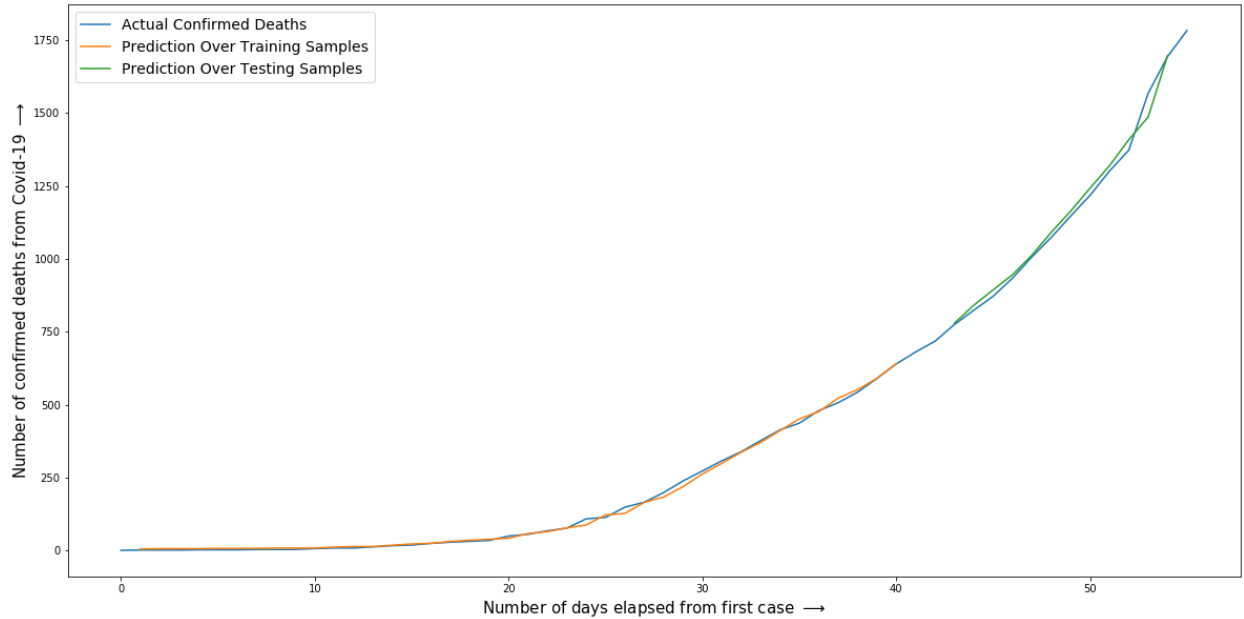

**Fig. 2.** The predicted trend of COVID-19 confirmed death cases from 1st confirmed case till 100 days

**Table 1.** Performance of the classifiers

| Regressor | Training Time (milliseconds) | Prediction Time (milliseconds) | Prediction Root Mean Squared Error (RMSE) | Lower Bound of Test Set | Upper Bound of Test Set |
|---|---|---|---|---|---|
| Bayesian Ridge (BRR) (Confirmed Cases) | 1.3067 | 0.1402 | 2031.08 | 9152 | 52952 |
| Gaussian Process (GPR) (Confirmed Cases) | 14.948 | 0.1478 | 2089.13 | 9152 | 52952 |
| Bayesian Ridge (BRR) (Death Cases) | 1.7235 | 0.3426 | 29.50 | 718 | 1783 |
| Gaussian Ridge (GPR) (Death Cases) | 62.226 | 0.4079 | 29.71 | 718 | 1783 |

From table 1, we can observe that to determine the performance of the regressors to predict the estimated confirmed COVID-19 positive cases, we can see that the root mean squared error (RMSE) for BRR is 2031.08 and GPR is 2089.13 where the lower and upper bounds of the test sets are 9152 and 52952 respectively. Now as the RMSE for both the regressors are very close, to test the performance of the classifiers, we need to check which one among them consumes lesser resources. For this, we can see that BRR trains the model in 1.3067 milliseconds (ms) and generates output in 0.1402 ms whereas GPR trains the model in 14.948 ms and generates output in 0.1478 ms. Both of the regressors' prediction times are very close but we can observe that BRR trains approximately 10 times faster than GPR. Based on all the parameters, we can confirm that BRR performs the best to determine the estimated confirmed COVID-19 positive cases based on currently available statistics.

Now for the death cases prediction, we can see that RMSE of BRR and GPR are 29.50 and 29.71 respectively for the test dataset having lower bound of 718 and upper bound of 1783. Now as the RMSE of both the classifiers is close, looking at the training and prediction time we can see that BRR trains 36 times faster and predicts 1.1 times faster than GPR. Therefore we see that in both the cases, BRR outperforms GPR in almost every aspect and we can reliably use BRR algorithm for the prediction of both COVID-19 confirmed cases as well as death cases caused by the virus.

**4.2 Performance analysis of COVID-19 detection with chest X-Rays model**

As discussed in section 3.3, a convolution neural network extracts several properties from the image to establish a classifiable relationship between the classes of images. Fig. 3 gives an overlook of the intermediate feature extraction process of CNN algorithm. A COVID-19 infected chest X-Ray was passed through the model, and CNN extracted several features from that image and those are reflected in each frame of fig. 3, this is how our model sees the image. After extraction of several such features, our model figured out the most important features from the dataset which can be used to classify the data. After training the model for several rounds until a consistent validation loss was obtained, our model was trained to classify COVID-19 from normal as well as lungs infected with other infections apart from COVID-19 by utilizing chest X-Ray images.

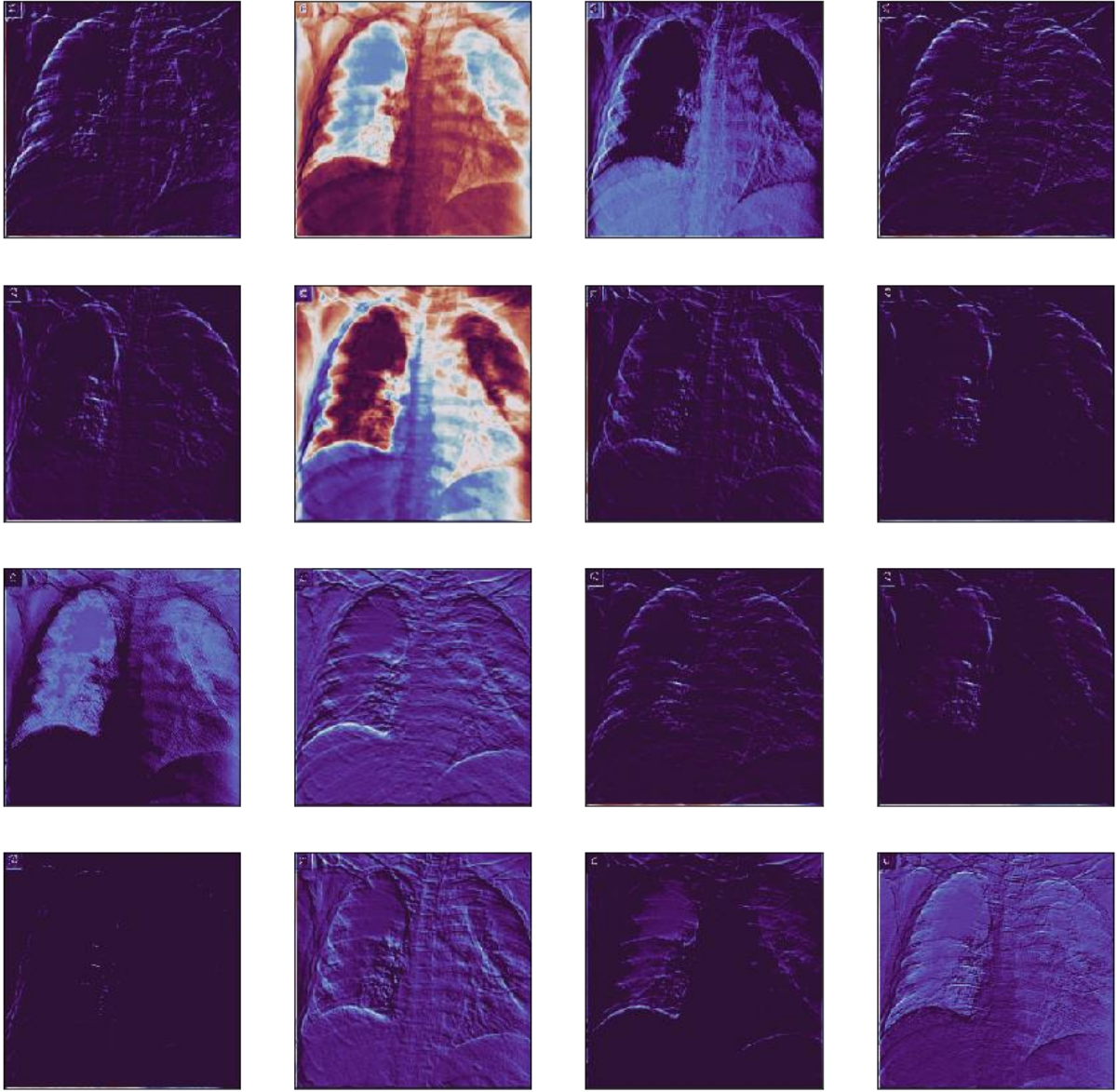

**Fig. 3.** How our deep learning model sees the COVID-19 infected chest X-Ray image to extract classifiable features

**Table 2.** The performance of the COVID-19 v/s Others classifier

|  | Accuracy (%) | Precision | Recall | F1 Score |
| --- | --- | --- | --- | --- |
| Performance | 95.02 | 0.944 | 1.000 | 0.968 |

From table 2, we can see that for the classification of the dataset, our model has obtained a classification accuracy of 95.02%. Now as the number of images in the two classes were different, therefore we use other metrics to understand the model performance at more depth. The precision and recall of the model are 0.944 and 1.000 respectively which means our model has not labelled any image as false-negative but few images were labelled as false-positive. The closer the score to 1, the better the model should perform. Same goes for F1 score which is the harmonic mean of precision and recall. The high value of accuracy, precision, recall and F1 score indicates that our model performed with high accuracy. Further development and testing with this model could be done to deploy as an inexpensive alternative COVID-19 testing strategy.

## 5 Conclusion

COVID-19 is the most serious disease outbreak of the 21$^{st}$ century and there is no specific cure available yet. To stop the spread of COVID-19, the best strategy is to prevent the community spread of the virus. To control the outbreak of the virus, science and technology have an immense role to play. Medical science is working very hard to discover a specific cure and other branches of science can also contribute to stop the spread of the virus. In this paper, we have discussed the role of Artificial Intelligence in controlling the pandemic. We have developed two strategies which can help in controlling the virus outbreak.

First, we have developed a machine learning algorithm with the help of Bayesian Ridge and Gaussian Process regressor techniques. We have trained it to predict the future trends of the virus outbreak based on currently available statistics. To estimated confirmed COVID-19 positive cases and death cases likely to happen in the near future, Bayesian Ridge Regressor performs the best in almost all aspect. By understanding the upcoming trends, it would be very useful to flatten the curve of reported incidence for a specific place.

After this, we have developed a deep learning model which could be useful to replace traditional tests. As the traditional test takes a long time and the testing centres are sparsely located, alternative testing strategies have high importance. With the help of our deep learning model, we have successfully classified Covid-19 positive patients from a dataset containing normal healthy individual and patients affected by other lung diseases by analyzing chest X-Ray images. For this purpose, our model has obtained a classification accuracy of 95.02%. This could help us to identify COVID-19 infected patients in a widely available and inexpensive manner.

Artificial intelligence (AI) could be a very useful tool to prevent the virus outbreak. Further development in AI could be done to predict the geographical expansion of the disease or aid officials to plan to serve an optimum amount of patients by monitoring the severity of their condition.

## 6 Code Availability

The code for the chest X-Rays classification model to detect COVID-19 has been made available in GitHub with MIT License for open source distribution and it can be freely reused for commercial or non-commercial purpose[25].